\documentclass{article}

     \usepackage[preprint]{neurips_2020}

\usepackage[utf8]{inputenc} 
\usepackage[T1]{fontenc}    
\usepackage{hyperref}       
\usepackage{url}            
\usepackage{booktabs}       
\usepackage{amsfonts}       
\usepackage{nicefrac}       
\usepackage{microtype}      
\usepackage{mathtools}
\usepackage{tikz}
\usepackage{graphicx}
\usepackage{pgfplots}
\usepackage{multirow}
\usepackage{subcaption}
\usepgfplotslibrary{groupplots}
\usepackage{natbib}
\usepackage{bm,bbm}
\usepackage[ruled,vlined]{algorithm2e}
\newcommand\R{\mathbb{R}}
\title{Shallow Neural Hawkes: Non-parametric kernel estimation for Hawkes processes}

\author{
  Sobin Joseph\\
  Department of Management Studies\\
  Indian Institute of Science\\
 Bangalore 560012\\
   \And
 Lekhapriya Dheeraj Kashyap\\
  Department of Management Studies\\
  Indian Institute of Science\\
 Bangalore 560012\\
 \And
 Shashi Jain\\
 Department of Management Studies\\
 Indian Institute of Science\\
 Bangalore 560012\\
  \texttt{shashijain@iisc.ac.in} \\
}

\begin{document}

\maketitle

\begin{abstract}
Multi-dimensional Hawkes process (MHP) is a class of self and mutually exciting point processes that find wide range of applications -- from prediction of earthquakes to modelling of order books in high frequency trading. This paper makes two major contributions, we first find an unbiased estimator for the log-likelihood estimator of the Hawkes process to enable efficient use of the stochastic gradient descent method for maximum likelihood estimation.  The second contribution is, we propose a specific single hidden layered neural network for the non-parametric estimation of the underlying kernels of the MHP. We evaluate the proposed model on both synthetic and real datasets, and find the method has comparable or better performance than existing estimation methods. The use of shallow neural network ensures that we do not compromise on the interpretability of the Hawkes model, while at the same time have the flexibility to estimate any non-standard Hawkes excitation kernel.
\end{abstract}

\section{Introduction}
Hawkes processes \citet{hawkes1971spectra} are temporal point processes in which the intensity depends on the process history with an excitation mechanism. It is well-known for studying seismic events \citet{ogata1999seismicity}, financial analysis \citet{filimonov2012quantifying}, \citet{bacry2015hawkes} and modelling social interactions \citet{crane2008robust}, \citet{blundell2012modelling}, \citet{zhou2013learning}. In the field of biology, it is used to study genomic events along DNA sequences \citet{reynaud2010adaptive}. MHP has also been used to model crime \citet{mohler2011self} and study the pattern of civilian deaths in Iraq \citet{lewis2012self}. The primary concern in modelling Hawkes process is the estimation of link function or excitation kernel. A common practice has been to assume a parametric form of the excitation kernel, the most common being exponential and power-law decay kernels, and then using maximum likelihood estimation  \citet{ozaki1979maximum} to determine the optimal values of the parameters.  

Formally, the multi-dimensional Hawkes process is defined by a $D$-dimensional point process $N_t^d\,\, ,d=1,\ldots,D,$ with the conditional intensity for the $d$-th dimension expressed as,

\begin{equation}
\lambda_d(t)=\mu_d + \sum_{j=1}^{D} \int_0^t \phi_{dj}(t-\tau)dN^j_{\tau},
\label{multidim_Hawkes}
\end{equation}

where $\mu_d$ is the exogenous base intensity for the $d$-th node and is  independent of the history.  $\phi_{dj},$  $1\leq d,j\leq D$ are called the excitation kernels that quantify the magnitude of excitation of the base intensity $\mu_d$ of the $d$-th node over time due to the past events from node $j.$ These kernel functions are positive and causal (their support is within $\mathbb{R}^+$). Inferring a Hawkes process requires estimating the base intensity $\mu_d$ and its kernels functions $\phi_{dj},$ either by assuming a parametric form for the kernels or in a non-parametric fashion. Recent developments focus on data-driven, non-parametric estimations of MHP to capture the general shape of the kernel and increase flexibility of the model.

In general, the kernels $\phi_{dj}$ as well as the base intensity $\mu_d$ can be estimated by maximizing the associated log-likelihood function. However,  as will be discussed further in Section \ref{prelims},  the challenge is that the log-likelihood function contains the integral of intensity function $\lambda_d(t)$ that depends on the values of the  kernels over the whole time interval. In this paper we present an unbiased estimator for the log-likelihood function of the MHP, which makes application of  SGD for maximum likelihood estimation straightforward. It should be noted that as the log-likelihood function for MHP is usually  non-convex in the parameter space, even for the basic exponential kernels, the SGD or other optimization methods do not guarantee a global maximum. However, in our experiments we observe that SGD, with ADAM, \citet{kingma2014adam}, used for the adaptive learning rates, with prescribed choice of initialization gets sufficiently close to the optimal parameters in few iterations.  

The main contribution of the paper is development of a feed-forward neural network based non-parametric approach to estimate the kernels of  MHP. Specifically each excitation kernel of the MHP is modelled as a separate feed-forward network with a single hidden layer. The weights of the different networks are coupled with each other in the likelihood function and the optimal weights are then determined using the batch SGD with objective to maximize the log likelihood. The choice of using a shallow --a single hidden layered-- neural network is to ensure that, on one hand a closed form for the time integrated value of the excitation kernels can be obtained, while at the same time --by the virtue of the universal approximation theorem-- have the ability to approximate any excitation kernel to arbitrary precision. In this paper we only consider the excitation effect of new arrivals, i.e. the output of the excitation kernel ranges in $\mathbb{R}^+,$  and  a fixed base intensity.  We test our model against few state-of-the-art non-parametric estimation methods for MHP. The method is tested against both synthetic as well as real data set. For real dataset we consider  the high frequency data of buy and sell market orders from Binance cryptoexchange. We find that the performance of our method, which we call the \emph{Shallow Neural Hawkes} (SNH) is comparable or better than benchmark models. We see a distinct advantage of our approach, in comparison to the use of recurrent neural networks to model MHP, as we do not lose the interpretability of MHP by recovering the underlying excitation kernels. Another advantage our method is that a closed form expression for the integrated kernel functions is obtained, which  --for instance-- histogram based non-parametric methods would require discrete time approximations.

\section{Related Work}

In many real world applications the flexibility of Hawkes process is enhanced by the use of non-parametric models. The first  non-parametric model of one dimensional Hawkes processes was proposed in \citet{lewis2011nonparametric} , based on ordinary differential equation (ODE).  The first extension of non-parametric kernels to multi-dimensional case was provided in \citet{zhou2013learning}. They developed an algorithm to learn the decay kernels by using Euler-Lagrange equations for optimization, in infinite dimensional functional space. Determined to model large amount of data, a non-parametric method based on solving the Wiener-Hopf equation using a Gaussian quadrature method was introduced in \citet{bacry2014second}. Motivated by the branching property of Hawkes process \citet{doi:10.1198/016214502760046925}, an Expectation-Maximization(EM) algorithm was developed in \citet{marsan2008extending} for non-parametric estimation of decay kernel and background intensity. 

The methods close to our approach include the MEMIP (Markovian Estimation of Mutually Interacting Processes) \citet{lemonnier2014nonparametric} that makes use of polynomial approximation theory and self concordant analysis to learn the kernels and  the base intensities. While the non-parametric models in (\citet{lemonnier2014nonparametric}; \citet{zhou2013learning}) represent excitation functions as a set of basis functions, a guidance for the selection process of basis functions  is provided  in \citet{xu2016learning}. Both \citet{xu2016learning} and \citet{salehi2019learning} express the excitation kernels as sum of Gaussian basis kernels, the former uses sparse group-lasso regularizer and is suitable for large datasets, while the latter uses variational expectation-maximization and is suitable for a handful of datasets. The approach presented in this paper is similar, as the excitation function is expressed as a non-parametric function,  specifically as exponential of sum of rectified linear units (ReLUs). 

In a relatively new study of temporal point processes, the authors in \citet{du2016recurrent} develop a recurrent neural network to model point processes and learn influences from event history. The authors in \citet{mei2017neural} develop a novel continuous-time LSTM to model self-modulating Hawkes processes. This setting can capture the exciting and inhibiting effects of past events on future and allow the background intensity to take negative values corresponding to delayed response or inertia of some events. Compared to the approach of expressing each excitation kernel as a  neural network,  LSTM might be less desirable when there is a greater focus of the interpretability of the MHP, for instance for learning the Granger causality graph. We also significantly simplify the SGD formulation as compared to  \cite{mei2017neural}, where one has to rely on simulations to obtain the gradients, while in \cite{du2016recurrent}  numerical integration is needed to obtain the necessary gradients of the log likelihood.

\section{Preliminary Definitions}\label{prelims}

A $D$-dimensional MHP is a collection of $D$ univariate counting processes $N_d(t)\, , d = 1,\ldots,D.$  The realization of MHP  over an observation period $[0, T)$ consists of a sequence of discrete events $\mathcal{S} = \{(t_n, d_n)\},$ where $t_n \in [0, T)$ is the timestamp of the $n$-th event and $d_n \in \{1,\ldots,D\}$ is the label of corresponding dimension in which the event occurred. The conditional intensity process for the $d-$th dimension is given by Equation \ref{multidim_Hawkes}.  Often the Hawkes kernels are assumed to be exponential function of the form $\phi_{dj}(t) = \alpha_{dj} e^{-\beta_{dj} t},$ and the base intensity $\mu_d$ is assumed to be constant. In this paper we assume that $\phi_{dj}:\mathbb{R} \rightarrow \mathbb{R}^+$ can be an arbitrary continuous function while $\mu_d$ is a positive constant. 

We denote the parameters of the multi-dimensional Hawkes process in a matrix form as $\bm{\mu} = [\mu_1,\ldots,\mu_D]^{\top}$ for the base intensity, and $\Phi = (\phi_{dj})$ for the excitation kernels. 
These parameters can be estimated by optimizing the log-likelihood over the observed events that are sampled from the process . The log-likelihood for model parameters $\Theta = \{\Phi, \bm{\mu}\}$ of Hawkes process can be derived from its intensity function (see for instance \citet{rubin1972regular},\citet{daley2007introduction}) and is given by,

\begin{eqnarray}\nonumber
\mathcal{L}(\Theta) &=& \sum_{d=1}^{D} \left( \int_0^T \log\left(\lambda_d(u)\right)\,dN_d(u) - \int_0^T \lambda_d(s)\,ds\right)\\
&=& \sum_{d=1}^{D} \left( \sum_{(t_n, d_n)\in \mathcal{S}}\left(\log\left(\lambda_d(t_n) \right) \mathbbm{1}{\{d_n=d\}}\right) - \int_0^T \lambda_d(s)\,ds\right)
\end{eqnarray}\label{logLikelihood}

For the application of SGD we need an unbiased estimator for the gradient of $\mathcal{L}$ with respect to model parameters. Obtaining an unbiased estimator for $\int_0^T \lambda_d(s)\,ds$ is challenging. \citet{mei2017neural} use a simulation based approach for an unbiased estimate, while \citet{yang2017online} work with a time-discretized version of $\mathcal{L}.$ Both these approaches are computationally intensive. We propose the following as an unbiased estimator for the gradient of the log likelihood function $\mathcal{L},$

\begin{equation}\label{unbiasedLogLikelihood}
\nabla_{\Theta}\left(\log(\lambda_{d_n}(t_n) ) - \int_{t_{n}^-}^{t_n} \mu_{d_n}\, ds -\sum_{j=1}^{D} \int_0^{T-t_n} \phi_{jd_n}(s)\,ds\right),
\end{equation}

where $(t_n,d_n) \in \mathcal{S}$ and $t_n^-:=\max\limits_{t_m}\{t_m | t_m < t_n \land d_m = d_n\},$ i.e. $t_n^-$ is the timestamp of the event that occurred just prior to the event at $t_n$ in node $d_n.$  The proof that the expression in Equation \ref{unbiasedLogLikelihood} is an unbiased estimator of the gradient of $\mathcal{L}$ is provided in Appendix \ref{A1}. A challenge in efficiently utilizing Equation \ref{unbiasedLogLikelihood} in the SGD method is that we need  a closed form expression for computing $\int \phi_{dj}(s)\,ds.$ When a parametric form for the excitation kernel is assumed, usually closed form expression for this integral exists. In Appendix \ref{B1} we present results for parameters inferred using SGD for exponential kernels and find that the results are close to the true parameter values. However, in the next section we present a non-parametric approach, which is general enough to infer any continuous excitation kernel, and also has closed form expression for the integrated excitation kernel. 

\section{Proposed Model}\label{model}

A feed-forward network with a single hidden layer, sufficiently large number of neurons, and  with appropriate choice of activation function is known to be a universal approximator \citet{hornik1989multilayer}.  We in the proposed method model each excitation kernel $\phi_{dj}(t),\, 1 \leq d,j \leq D$ of the MHP using a separate feed-forward network with a single hidden layer. As we consider only excitation kernels, the output of each of these neural networks should be in $\mathbb{R}^+.$ The weights of the different networks are coupled with each other in the likelihood function. We use the batch stochastic gradient descent to maximize the log likelihood over the parameter space, where the unbiased estimates of the gradient of the log-likelihood  are obtained using Equation \ref{unbiasedLogLikelihood}. For efficient calculation of the gradient, as discussed in Section \ref{prelims} ideally there should be a closed form expression for the time integrated value of the approximated excitation kernel. Based on these criterion , a positive output for the approximated excitation kernel and its integral with a closed form expression, we came up with a specific architecture for our neural network.

In order to approximate $\phi_{dj}(t),\, 1 \leq d,j \leq D$ we use a feed-forward network $\widehat{\phi}_{dj} : \mathbb{R} \rightarrow \mathbb{R}^+$ of the form

\begin{equation*}
\widehat{\phi}_{dj} := \psi \circ A_2 \circ \varphi \circ A_1
\end{equation*}
where $ A_1:\R \rightarrow \R^p$ and $ A_2:\R^p \rightarrow \R$ are affine functions of the form,

\begin{equation*}
A_1(x) = \mathbf{W}_1 x + \mathbf{b}_1 \ \ \textrm{for} \ x \in \R,\ \mathbf{W}_1 \in \R^{p \times 1}, \mathbf{b}_1 \in \R^p,
\end{equation*}
and
\begin{equation*}
A_2(\mathbf{x}) = \mathbf{W}_2\mathbf{x} + b_2 \ \ \textrm{for} \ \mathbf{x} \in \R^p,\ \mathbf{W}_2 \in \R^{1 \times p}, b_2 \in \R.
\end{equation*}

$\varphi : \R^j \rightarrow \R^j, j \in \mathbb{N}$ is the component-wise ReLU activation function given by:
\begin{equation*}
\varphi(x_1,\ldots,x_j):=\left(\max(x_1,0),\ldots,\max(x_j,0)\right),
\end{equation*}

while $\psi:\R \rightarrow \R+,$ is exponential function
\begin{equation*}
\psi(x):= e^{x}
\end{equation*}

With a choice of $p$ neurons for the hidden layer, the dimension of the parameter space for the network will be $3p+1.$ For a $D$-dimensional Hawkes process we would need $D^2$ networks. Writing $\mathbf{W}_1 := [\beta_1,\ldots,\beta_p]^{\top},$ $\mathbf{W}_2 := [\alpha_1,\ldots,\alpha_p],$ the approximate kernel can be written as:

$$
\widehat{\phi}_{dj}(x) = \exp{\left(b_2 + \sum_{i=1}^p \alpha_i \max\left(\beta_i x + b_1^i,0\right)\right)}
$$ 

The choice of exponential function for the output layer is to ensure that the output is in $\R^+$ as required by excitation kernels.  As ReLU activation function is not a polynomial everywhere, the network will be a universal approximator (\citet{leshno1993multilayer}). The other advantage of this particular choice of network architecture is that a closed form expression for $\int_0^t \widehat{\phi}_{dj}(u)\,du$ can be readily evaluated, and turns out to that it is  a linear combination of $\widehat{\phi}_{dj},$ see Appendix \ref{A2} for details. The optimal parameters for the MHP, i.e. $\Theta = \{\Phi, \bm{\mu}\},$ where $\Phi$ is the set of weights of all the $D^2$  networks, is obtained using batch SGD, where we use ADAM for the adaptive learning rates.

\section{Experiments and Results}
\subsection{Synthetic Data}
In this section we demonstrate the performance of the Shallow Neural Hawkes model by fitting various forms of kernels and by weighing it against state-of-the-art non-parametric models, including EM method given in \citet{lewis2011nonparametric} and Wiener-Hopf (WH) model described in \citet{bacry2014second}. All simulations are performed using the thinning algorithm described in \citet{ogata1981lewis}. We also use large set of tools from the tick library, \citet{bacry2017tick}, that facilitates efficient parametric and non-parametric estimations. Here we examine the univariate case of Hawkes process, followed by the bivariate case.

\subsubsection{Univariate Case} \label{univariate}
First, we simulate univariate Hawkes process for widely used forms of kernels, namely 
\begin{equation}	
\text{Exponential kernel: } ~~~\phi(t) = \alpha e^{-\beta t}
\end{equation}  
\begin{equation}
\text{Power Law kernel: }~~~ \phi(t) = \alpha (\delta+t)^{-\beta}
\end{equation}  

Next, we verify the performance of the Shallow Neural Hawkes model on a rectangular kernel given by,
\begin{equation}
\phi(t) = \begin{cases}
              \alpha \beta, & \text{if}\ \delta < t < \delta + \dfrac{1}{\beta} \\
              0, & \text{otherwise}
              \end{cases}
\end{equation}  

\textbf{Experiment setup :} 
For the exponential kernel simulation, we use parameters $[ \alpha, \beta, \mu] = [1,4,0.05]$, for a period of $[0,60000)$ and we get $N_T = 3972$ events. Similarly, for the simulation of power law kernel, we use parameters $[ \alpha, \beta, \delta, \mu] = [1,4,1,0.05]$, for a period of $[0,60000]$ and get $N_T = 4442$ events. We use  $100$ neurons for each kernel and the initial weights are drawn from uniform distribution in the range of $[0,0.5].$ In all our initializations we find that positive weights for the inner layer and negative weights for the outer layer helps in faster convergence of the algorithm. This initial setting is common to all experiments in this paper. We use ADAM optimizer \citet{kingma2014adam}, set the batch size to 50 and employ varied learning rates for the parameters of the inner and the outer layer. We find by default using a learning rate of  $10^{-2}$ for the outer layer and $10^{-5}$ for the inner layer, and $10^{-3}$ for $\mu$ gives reasonably good results. We train the network up to 30 epochs. 

The rectangular kernels are simulated using parameters $[ \alpha, \beta, \delta, \mu] = [0.7,0.4,1,0.05]$, for a period of $[0,60000]$ and we get $N_T = 10196$ events. Implementation details for the SNH model is similar to the above setting. The learning rate for outer layer is $10^{-2}$, for inner layer is $5 \times 10^{-4}$ and for $\mu$ is $10^{-3}$. The model is trained for 30 epochs.  When the excitation kernels are smooth we find using smaller learning rates for the inner layer can significantly improve the convergence, although for kernels with inflection points (like the rectangular kernel) a relatively higher learning rate for the inner layer helps. For all the cases we find that using a higher learning rate for the output layer in comparison to the learning rate for the input layer helps in faster convergence. 

\begin{figure}
  \centering
  \begin{subfigure}[b]{0.3\linewidth}
    \includegraphics[width=\linewidth]{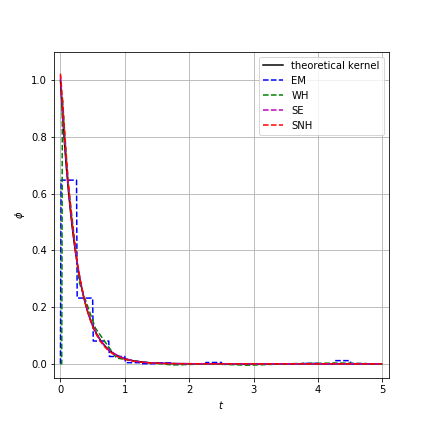}
    \caption{}
  \end{subfigure}
  \begin{subfigure}[b]{0.3\linewidth}
    \includegraphics[width=\linewidth]{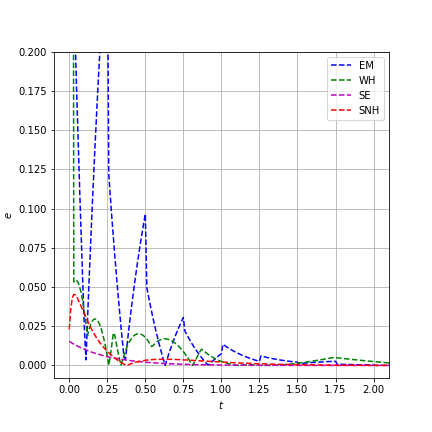}
    \caption{}
  \end{subfigure}
  \begin{subfigure}[b]{0.3\linewidth}
    \includegraphics[width=\linewidth]{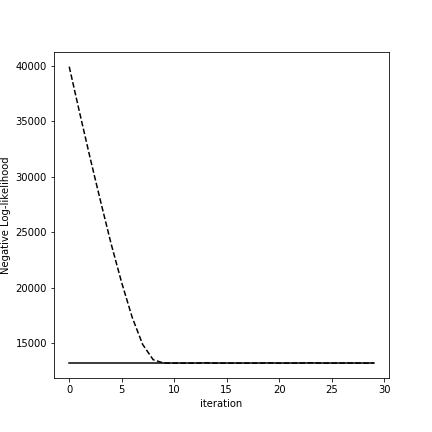}
    \caption{}
  \end{subfigure}
  \begin{subfigure}[b]{0.3\linewidth}
    \includegraphics[width=\linewidth]{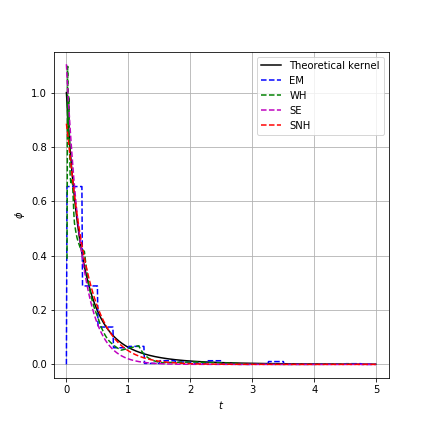}
    \caption{}
  \end{subfigure}
  \begin{subfigure}[b]{0.3\linewidth}
    \includegraphics[width=\linewidth]{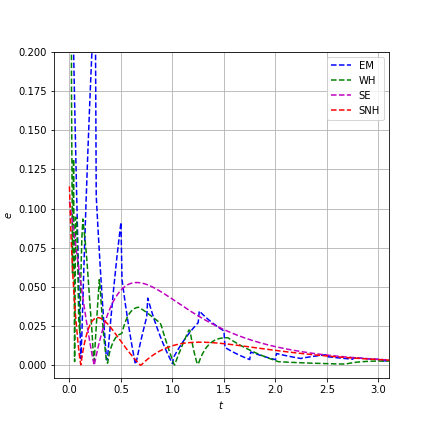}
    \caption{}
  \end{subfigure}
  \begin{subfigure}[b]{0.3\linewidth}
    \includegraphics[width=\linewidth]{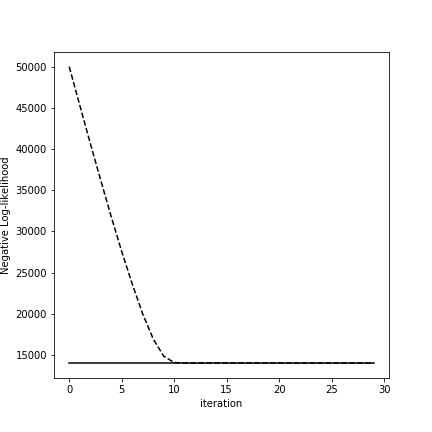}
    \caption{}
  \end{subfigure}
  \begin{subfigure}[b]{0.3\linewidth}
    \includegraphics[width=\linewidth]{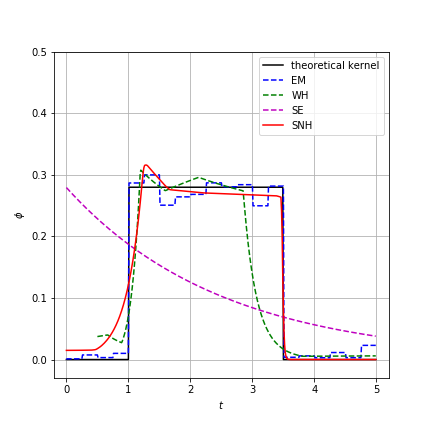}
    \caption{}
  \end{subfigure}
  \begin{subfigure}[b]{0.3\linewidth}
    \includegraphics[width=\linewidth]{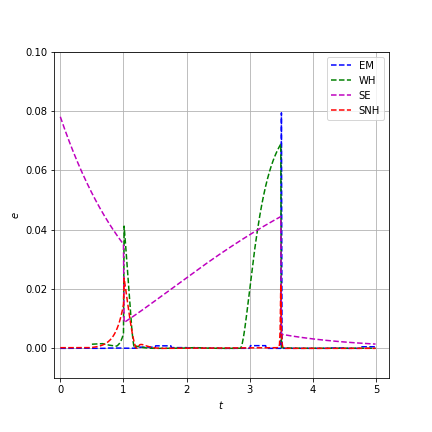}
    \caption{}
  \end{subfigure}
  \begin{subfigure}[b]{0.3\linewidth}
    \includegraphics[width=\linewidth]{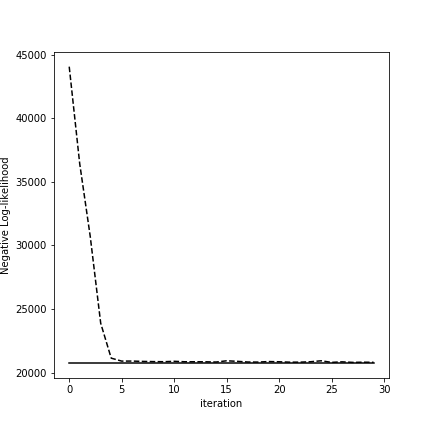}
    \caption{}
  \end{subfigure}
  \caption{\footnotesize Synthetic data experiment results for univariate Hawkes processes \\ (a) The estimated Exponential kernel  (b) The estimated error for Exponential kernel  (c) The convergence plot of Negative Log-likelihood for Exponential kernel estimation in SNH model  (d) The estimated Power Law kernel  (e) The estimated error for Power Law kernel  (f) The convergence plot of Negative Log-likelihood for Power Law kernel estimation in SNH model  (g) The estimated rectangular kernel  (h) The estimated error for rectangular kernel  (i) The convergence plot of Negative Log-likelihood for rectangular kernel estimation in SNH model}
  \label{Synthetic_data}
\end{figure}

The learned kernels from the Shallow Neural Hawkes model are then compared to kernels determined by the parametric sum of exponential kernels method, non-parametric EM and WH model, as shown in Figure \ref{Synthetic_data}. These models are implemented using the tools provided in tick library \citet{bacry2017tick}.For the non-parametric EM estimation, we choose the kernel support as $5$ and kernel size of $20$. For the WH method, we set the number of quadratures as $50$ and use linear sampling for exponential kernels. However, the linear sampling method performs poorly in the case of power law kernel and rectangular kernel, hence we use the semi-log sampling approach with maximum kernel support of $1000$ and maximum lag as $100$. The next section provides a detailed description of the observations.

\textbf{Experiment results :} 
First, we compare the performance of non-parametric models based on the kernel estimation approach. The EM model is a histogram based estimator with discrete function kernel, whose performance critically depends on the choice of bins. WH model also has a strong dependancy on the choice of grid in the kernel estimation process \citet{morzywolek2015non} . On the contrary, the Shallow Neural Hawkes model provides a continuous function kernel and does not rely on the range of kernels, a vital advantage of the model. 

From a visual assessment of the kernel estimation plot in \ref{Synthetic_data} , it is evident that the Shallow Neural Hawkes model outperforms the former models in the exponential and power law kernel estimation and exhibits a finer performance in the case of the rectangle kernel. Next on the evaluation metrics, we plot the L1 error, defined as $|\phi - \phi_{\text{est}}|$, between the true and estimated kernels of all the models in comparison, refer Figure\ref{Synthetic_data}. We observe that the error is consistently significantly lower in the case of Shallow Neural Hawkes. Figure\ref{Synthetic_data} also shows the convergence of the system, the negative log-likelihood estimated in the Shallow Neural Hawkes model reaches the true negative log-likelihood precisely within 10 epochs. We have verified that speed of convergence is higher for larger sample periods. 

\subsubsection{Bivariate Case}
We simulate bivariate Hawkes processes for the exponential and power-law kernels using tick library, the complete experimental setup and results are discussed in the Appendix \ref{B2}. In this section, we define some random kernels to test the performance of the Shallow Neural Hawkes model, this helps us understand the versatility of our model.

\textbf{Experiment setup :} The random kernels are simulated using the TimeFunction class from tick library, it uses several types of interpolation to determine the function value between two points on $[0,\infty)$ \citet{bacry2017tick}. The kernel function $\phi_{(0,0)}(t)$ is defined using $x = [0, 1, 1.5, 2., 3.5],  y = [0, 0.2, 0, 0.1, 0.]$ and the y-values are extended to the right. Next, we have $\phi_{(0,1)}(t) = \dfrac{sin(t)}{4} ~~\text{for} ~0 < t  < T$. We then generate a zero kernel $\phi_{(1,0)}(t)$ = 0. Finally, we simulate a random form kernel for $\phi_{(1,1)}(t)$ using $x = [0., .7, 2.5, 3., 4.]$ and $y =  [.3, .03, .03, .2, 0.]$. The baseline values are set at $\mu = [0.05, 0.05]$. To recover the random kernels using SNH model, we use the same network setting as above with the learning rate for the inner layer set to $5 \times 10^{-4}$. We train the network for 100 epochs.

\begin{figure}
  \centering
  \begin{subfigure}[b]{0.4\linewidth}
    \includegraphics[width=\linewidth]{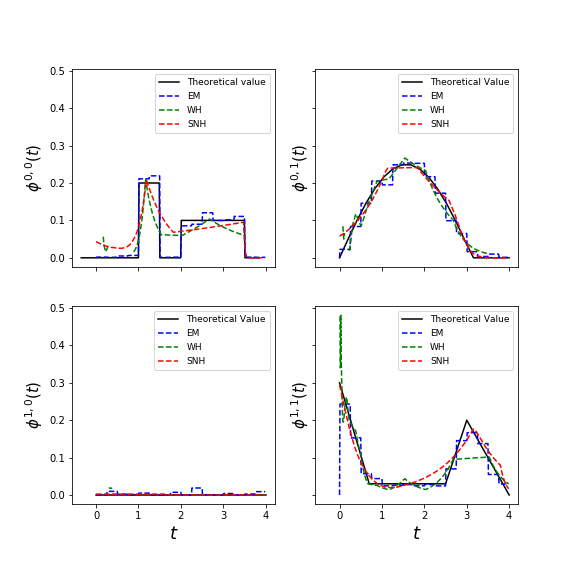}
    \caption{}
  \end{subfigure}
  \begin{subfigure}[b]{0.4\linewidth}
    \includegraphics[width=\linewidth]{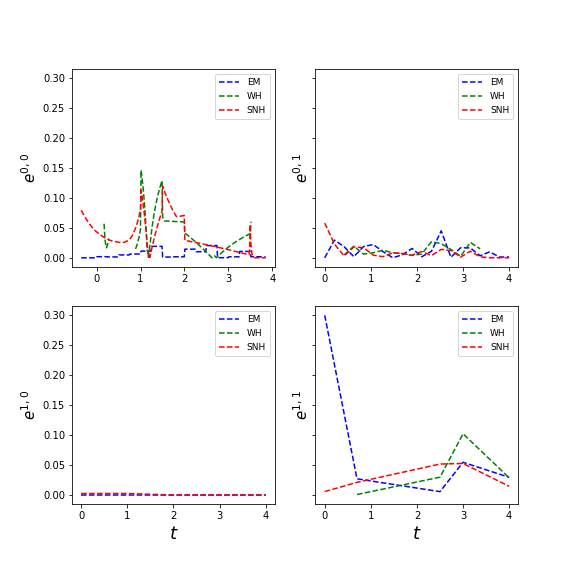}
    \caption{}
  \end{subfigure}
  \caption{\footnotesize Synthetic data experiment results for bivariate Hawkes processes with random kernels \\ (a) The estimated random kernels  (b) The estimated error for random kernels}
  \label{Synthetic_data_2}
\end{figure}

\textbf{Experiment results :} Here we discuss the performance of the SNH model on random kernels, as shown in Figure \ref{Synthetic_data_2}. We see that the kernel setting in $\phi_{0,0}(t)$ is highly disadvantageous to the SNH model. However, the EM model exhibits an impressive performance in fitting $\phi_{0,0}(t)$, while the WH struggles to capture this function. In the case of kernels  $\phi_{0,1}(t) , \phi_{1,0}(t), \phi_{1,1}(t)$, the SNH model achieve better results when compared to other models and this serves as proof that our can be applicable to a diverse class of non-parametric Hawkes processes.

Finally, we have verified that the SNH model accurately recovers the baseline values in both, univariate and bivariate case of Hawkes processes.

\subsection{Real Data}
A growing literature is dedicated to study the application of point processes to high frequency financial data. In particular, due to the correlated and clustered nature of trading activity, Hawkes processes are used to model trade arrival dynamics. A continuous time bivariate Hawkes process was used for modelling the arrival times of market sell and buy orders in \citet{bowsher2007modelling}. Recently, a bivariate Hawkes process was proposed in \citet{bacry2013modelling} to model the variations of asset prices and study the signature plot and the Epps effect. In this paper, we investigate the performance of the Shallow Neural Hawkes model on arrival data for buy and sell bitcoin market orders on the Binance exchange.

\textbf{Experiment setup :} We use the bitcoin data, traded in the Binance cryptocurrency exchange. The full dataset consists of 120000 intraday market orders, as recorded on 08 May 2020 covering the period between 6.45 PM to 10.55 PM (UTC), with corresponding volume and timestamps rounded to nearest second. The dataset was cleaned to include only unique market orders, as a particular market order might require several limit orders to full fill the demanded volume; with each recorded as a separate trade with a common market order id.

A bivariate analysis is performed jointly on the buy and sell trade data, to learn the interactions between them. For the SNH network architecture, we use the same initial settings as in the synthetic data instance. We set a learning rate of  $10^{-2}$ for the outer layer,  for the inner layer we use a learning rate of $10^{-3},$ and for $\mu$ a learning rate of $10^{-3}$. With $N_{T_{\text{sell}}} + N_{T_{\text{buy}}} = 83574$, we train the network in 30 epochs. 

To facilitate comparison with standard models, we perform non-parametric analysis on the bitcoin dataset using EM and WH models. For the EM estimation, we choose the kernel support as $6$ and kernel size of $100$. For the WH method, we set the number of quadratures as $200$.

\begin{figure}
 
  \centering
  \includegraphics[width=\linewidth]{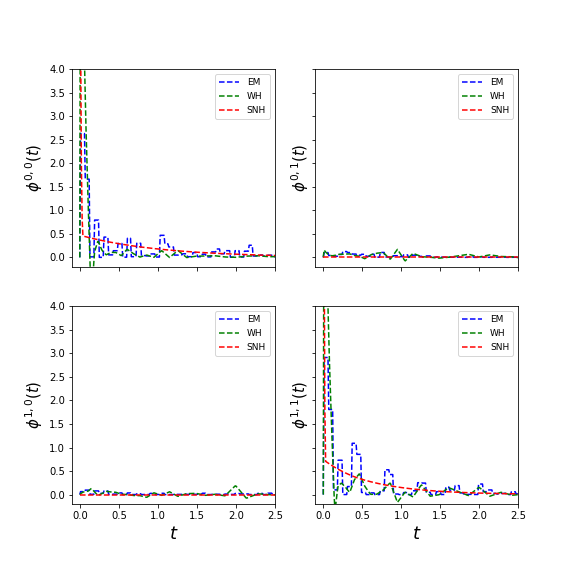}
  \caption{\footnotesize Experiment results of Bitcoin data as bivariate Hawkes processes. Estimated $\mu_1$ is 0.857, 0.463, 0.070 for SNH, EM and WH respectively. Estimated $\mu_2$ is 1.173, 0.685, 0.272 for SNH, EM and WH respectively. The negative log-likelihood values are -40143,-33127, -29698 for SNH, EM and WH respectively. }
  \label{Real_data_3}
  
\end{figure}

\textbf{Experiment results :}  In Figure \ref{Real_data_3}, we plot the kernels estimated by SNH, EM and WH methods. It is evident that the two events are not mutually exciting, but exhibit self exciting behaviour. The negative log-likelihood values recorded from the SNH, EM and WH models are -40143, -33127, and -29698 respectively. The SNH model achieves competitive negative loglikelihood when compared to EM and WH models. We also perform a k-fold cross validation to compare the three models on limited samples of data, with negative log-likelihood used as an evaluation metric (see Appendix \ref{B3} for more details). The WH method exhibits consistently poor results while the performances of EM and SNH methods are comparable.

\section{Conclusion}
We have developed a non-parametric kernel estimation method for the MHP, which we call the Shallow Neural Hawkes. The SNH models the excitation kernel as a feed-forward network with a single hidden layer. To ensure that we can efficiently determine the optimal parameters using the SGD, and that the kernels are excitation kernels, we arrive at a specific architecture for the network.  The excitation kernel then translates to an exponential of sum of ReLU functions. The parameters of the network are obtained using a batch SGD with log-likelihood as the objective to maximize. We  provide an unbiased estimator for the gradient of the log-likelihood function required for efficient application of SGD. The method is tested with both synthetic and real data set. The real data set consists of tick-by-tick buy and sell market orders for bit-coins on binance crypto-currency exchange. The performance of our method is consistently comparable with the best in all the examples considered.

\medskip

\small
\bibliographystyle{plainnat}
\bibliography{notes_1}
\normalsize
\newpage 
\appendix
\section{Derivation of Expressions}
\subsection{Unbiased gradient estimator for the log likelihood function of MHP}\label{A1}

We want to determine an unbiased estimator for the gradient of the log-likelihood function for MHP.
\begin{eqnarray}\nonumber
\mathcal{L}(\Theta) &=& \sum_{d=1}^{D} \left( \int_0^T \log\left(\lambda_d(u)\right)\,dN_d(u) - \int_0^T \lambda_d(s)\,ds\right)\\
&=& \sum_{d=1}^{D} \left( \sum_{(t_n, d_n)\in \mathcal{S}}\left(\log\left(\lambda_d(t_n) \right) \mathbbm{1}{\{d_n=d\}}\right) - \int_0^T \lambda_d(s)\,ds\right)\label{int0}
\end{eqnarray}

Let $\{t_1^d,\ldots, t_{N_d^T}^d\},$ be the ordered arrival times for the nodes $d=1,\ldots, D.$ We first  focus on the integral of the intensity with respect to time. 

\begin{eqnarray}
\int_0^T \lambda_d(s)\,ds &=& \int_0^T \left(\mu_d + \sum_{t_n<s} \phi_{dd_n} (s-t_n)\right) ds \\
&=& \int_0^T \mu_d \, ds + \int_0^T \left( \sum_{t_n<s} \phi_{dd_n} (s-t_n)\right) ds \label{int1}
\end{eqnarray}

We can write the first part of the integral as 
\begin{equation}\label{int2}
\int_0^T \mu_d \,ds = \sum_{t_n^d<T} \int_{t_{n-1}^d}^{t_n^d} \mu_d\,ds
\end{equation}

The second part of the expression in Equation \ref{int1} can be written as follows 

\begin{eqnarray}
\int_0^T\sum_{t_n<s} \phi_{dd_n} (s-t_n)\, ds &=& \sum_{(t_n,d_n) \in \mathcal{S}} \int_{t_n}^{t_{n+1}} \sum_{t_m<t_n} \phi_{dd_m}(s-t_m) \,ds ,\\
&=&  \sum_{(t_n,d_n) \in \mathcal{S}} \int_{t_n}^T \phi_{dd_n}(s-t_n)\,ds, \\
&=& \sum_{(t_n,d_n) \in \mathcal{S}} \int_0^{T-t_n} \phi_{dd_n}(s)\,ds,\\
&=& \sum_{j=1}^D \sum_{t_i^j <T} \int_0^{T-t_i^j}\phi_{dj}(s)\, ds
\end{eqnarray}

where the first equality is from partitioning the interval $[0,T)$ by the arrival times, the  second equality comes from the fact that the term $\phi_{dd_n}(s-t_n)$ will appear in all integral partitions greater than $t_n,$ while the third equality is obtained by a basic change of variable. The final equality is a basic rearrangement of terms. 

Finally, we use the following relation obtained from the rearrangement of  the terms
\begin{equation}\label{int3}
\sum_{d=1}^D \sum_{j=1}^D \sum_{t_i^j <T} \int_0^{T-t_i^j}\phi_{dj}(s)\, ds = \sum_{d=1}^D \sum_{j=1}^D \sum_{t_i^d <T} \int_0^{T-t_i^d}\phi_{jd}(s)\, ds
\end{equation}

Substituting Equation \ref{int3} and \ref{int2} into Equation \ref{int0} gives us:

$$
\mathcal{L}(\Theta) = \sum_{d=1}^{D} \left( \sum_{t_n^d < T }\left(\log\left(\lambda_d(t^d_n) \right)  - \int_{t_{n-1}^d}^{t_n^d} \mu_d\,ds - \sum_{j=1}^D \int_0^{T-t_n^d}\phi_{jd}(s)\, ds \right)\right)
$$

Therefore gradient of $\mathcal{L}$ is:

$$
\nabla_{\Theta} \mathcal{L}(\Theta) = \sum_{d=1}^{D} \left( \sum_{t_n^d < T }\nabla_{\Theta} \left(\log\left(\lambda_d(t^d_n) \right)  - \int_{t_{n-1}^d}^{t_n^d} \mu_d\,ds - \sum_{j=1}^D \int_0^{T-t_n^d}\phi_{jd}(s)\, ds \right)\right),
$$

which gives us the unbiased estimator of Equation \ref{unbiasedLogLikelihood}.

\subsection{Integrated shallow excitation kernel}\label{A2}

As described in Section \ref{model} SNH models each excitation kernel $\phi_{dj}(t),$ as 

$$
\widehat{\phi}_{dj}(t) = \exp{\left(b_2 + \sum_{i=1}^p \alpha_i \max\left(\beta_i x + b_1^i,0\right)\right)},
$$ 

where $p$ is the number of neurons used in the hidden layer. The unbiased estimator in Equation \ref{unbiasedLogLikelihood} requires us to compute the gradient of the integrated excitation kernel, i.e.

$$
\int_0^t \phi_{dj}(s)\,ds.
$$

We here provide the expression for the integrated $\widehat{\phi}_{dj}(t).$ Let $\{s_1 \leq s_2 \leq \cdots \leq s_p\}$ be the sorted set inflection points for the $p$ neurons, where we define the inflection point of the $i$th neuron as,

$$
x_i = -\frac{b_1^i}{\beta_i}.
$$ 

Let $0\leq s_{l} \leq \cdots \leq s_ {u} \leq T, $ where $1 \leq l \leq u \leq p$ be the largest subsequence of the sorted inflection points, i.e. all the inflection points that lie in the range $[0,T].$ Then,

\begin{equation}\label{int4}
\int_0^t \widehat{\phi}_{dj}(s)\, ds = \int_0^{s_l} \widehat{\phi}_{dj}(s)\,ds+\cdots+\int_{s_u}^{T} \widehat{\phi}_{dj}(s)\,ds
\end{equation}

Equation \ref{int4} can be easily solved, as between two consecutive sorted inflection points , $0<s_m<s_n<T,$

$$
\int_{s_m}^{s_n} \widehat{\phi}_{dj}(s)\, ds = \frac{1}{\sum_{i=1}^p \alpha_i\beta_i \mathbbm{1}\left\{\lim_{x \to s_n^-} \beta_i x + b_1^i >0 \right\}} \left(\widehat{\phi}_{dj}(s_n)-\widehat{\phi}_{dj}(s_m)\right)
$$

\section{Additional Results}
\subsection{Parameter estimation for MHP using SGD}\label{B1}
We here report the results of parameter estimation for Hawkes processes with exponential kernel using the batch SGD where the gradient is computed using the unbiased estimator described in Equation \ref{unbiasedLogLikelihood}. As the log-likelihood function is non-convex in the parameter space for the exponential kernel, most common methods fix the value of decay $\beta$ and optimize upon the adjacency $\alpha.$ We find that with batch SGD, with ADAM used for adaptive learning rates we get fairly good results in few iterations.  Table \ref{synthetic_univariate_SGD} shows  the parameters estimated using SGD from simulated timestamps for different choices of true  parameter values of a one dimensional exponential Hawkes process. The simulation is done with $T$ set to  5000. We use a  learning rate of 0.01, a batch size of 32, and parameter values initialized using uniform random between 0 and 1. 

Table \ref{synthetic_multivariate_SGD} shows the parameter values estimated for a bivariate Hawkes Process. The choice of hyper-parameters are the same as that for the 1-D case. 

\begin{table}
	\centering
	\begin{tabular}{cccc}
		\hline
		$\mu$ & $\alpha$& $\beta$  &Actual Parameters \\
		(se)&(se)&(se)&\\
		\hline
		1.012&0.498&2.005&[1,0.5,2]\\
		(0.044)&(0.0211)&(0.208)&
		\\
		\hline
		2.05&3.07&9.57 &[2,3,10]\\
		(0.051)	&(0.037)&(0.32)&\\
		\hline
		0.489&186.1&585.38 & [0.5,200,600]\\
		(0.0128)&(5.68)&(10.24)&\\
		\hline
			
	\end{tabular}
	\caption{Estimated mean values of $\mu$, $\alpha$ and $\beta$ for univariate Hawkes simulated data.T The actual parameters are in order [$\mu$, $\alpha$, $\beta$ ]}\label{synthetic_univariate_SGD}
\end{table}

\begin{table}[!h]
	\centering
	\begin{tabular}{ccc}
		\hline
		$\mu$& $\alpha$& $\beta$  \\
		\hline
		$\begin{bmatrix}
		0.514 \\ 0.528\end{bmatrix}$&$\begin{bmatrix} 185.03 & 192.35\\ 186.48 & 195.09 \end{bmatrix}$& $\begin{bmatrix} 589.24& 561.29\\ 555.92 & 572.37\end{bmatrix} $ \\
		
		\hline
		
	\end{tabular}
	\caption{Estimated  $\mu,$ $\alpha,$ and $\beta$ matrix for exponential MHP. The actual parameters $\mu_d=0.5,$ $\alpha_{dj} =200,$ $\beta_{dj} = 600,$ where $1 \leq d,j \leq 2.$ }
	\label{synthetic_multivariate_SGD}
\end{table}
\subsection{Bivariate analysis of synthetic data}\label{B2}

In this section, we simulate bivariate Hawkes processes for the following kernels using tick library,
\begin{equation}	
\text{Exponential kernel: } ~~~\phi_{ij}(t) = \alpha_{ij} e^{-\beta_{ij} t}
\end{equation}  
\begin{equation}
\text{Power Law kernel: }~~~ \phi_{ij}(t) = \alpha_{ij} (\delta+t)^{-\beta_{ij}}
\end{equation}  

\textbf{Experiment setup :} We simulate the exponential kernel using $\alpha_{ij} = [[.3, 0.], [.6, .21]], ~\beta_{ij} = [[4., 1.], [2., 2.]], ~\mu = [0.12, 0.07]$ in the sample period $[0,5000]$ and $N_T = 1900$. For the power kernel simulation, we use  $\alpha_{ij} = [[1, 0.1], [0.6, 0.21]], ~\beta_{ij} = [[4., 4.], [4., 4.]], ~\delta = 1,  ~\mu = [0.05, 0.05]$ in the sample period $[0,50000]$ and $N_T= 7450$. For the SNH network architecture, we use the same initial settings as in the univariate instance. We also use the same optimising technique, batch size and learning rates for the hyperparameters. We train the network in 12 epochs with 38 randomly sampled batches (in the exponential kernel training) and 149 randomly sampled batches(in the case of power law kernel training) in one epoch, and verify the model at each epoch.

\begin{figure}
  \centering
  \begin{subfigure}[b]{0.4\linewidth}
    \includegraphics[width=\linewidth]{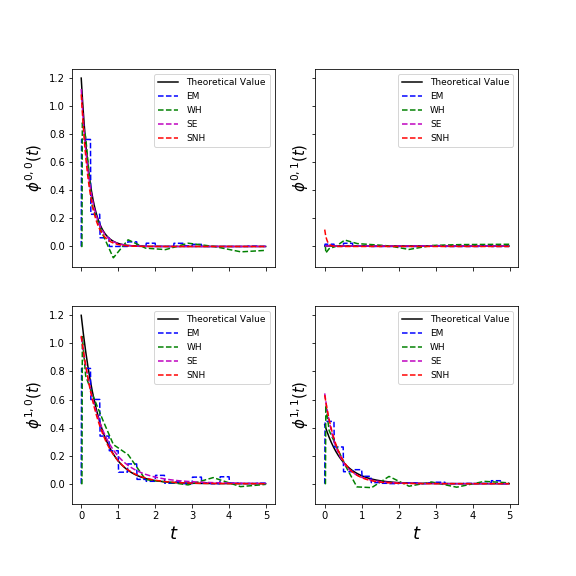}
    \caption{}
  \end{subfigure}
  \begin{subfigure}[b]{0.4\linewidth}
    \includegraphics[width=\linewidth]{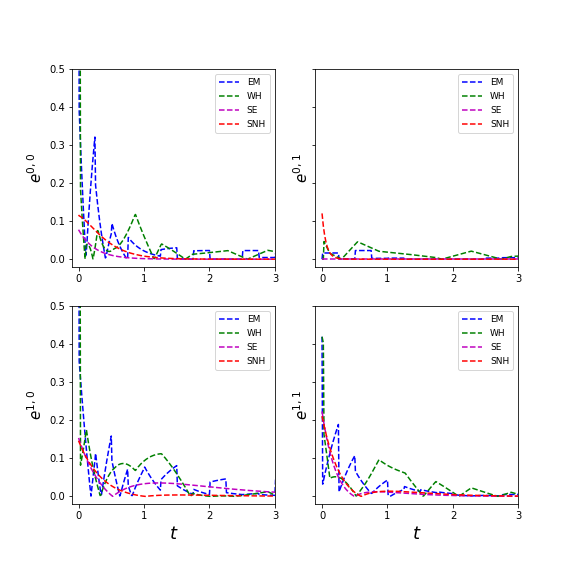}
    \caption{}
  \end{subfigure}
  \begin{subfigure}[b]{0.4\linewidth}
    \includegraphics[width=\linewidth]{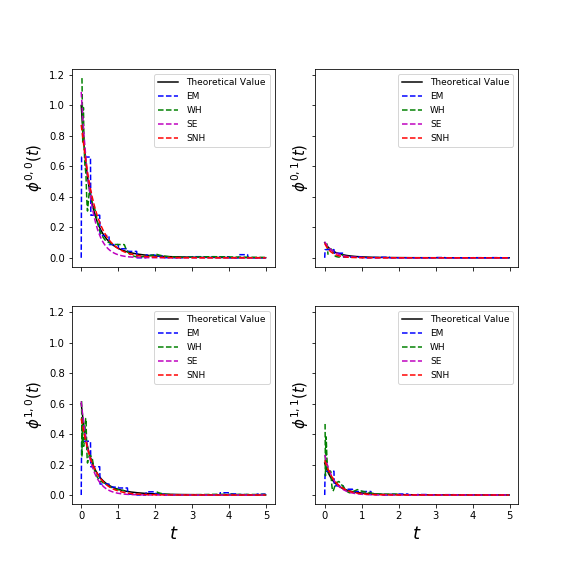}
    \caption{}
  \end{subfigure}
  \begin{subfigure}[b]{0.4\linewidth}
    \includegraphics[width=\linewidth]{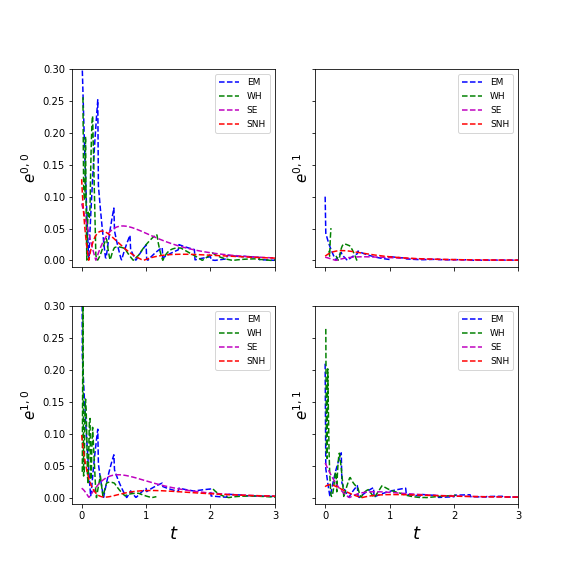}
    \caption{}
  \end{subfigure}
    \begin{subfigure}[b]{0.4\linewidth}
    \includegraphics[width=\linewidth]{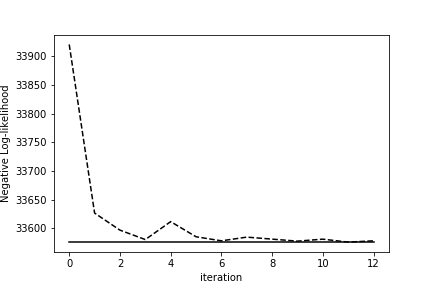}
    \caption{}
  \end{subfigure}
    \begin{subfigure}[b]{0.4\linewidth}
    \includegraphics[width=\linewidth]{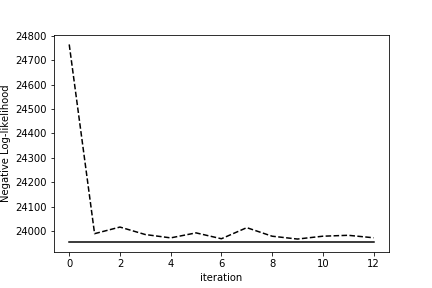}
    \caption{}
  \end{subfigure}
  \caption{Synthetic data experiment results for bivariate Hawkes processes \\ (a) The estimated Exponential kernels  (b) The estimated error for Exponential kernels  (c) The estimated Power Law kernels  (d) The estimated error for Power Law kernels  (e) The convergence plot of Negative Log-likelihood for Exponential kernels estimation in SNH model (f) The convergence plot of Negative Log-likelihood for Power Law kernels estimation in SNH model }
  \label{Synthetic_data_1}
\end{figure}

\textbf{Experiment results :} The figure  \ref{Synthetic_data_1} represents the kernels estimated by SNH model while we weigh it against the kernels generated by parametric sum of exponential model, non-parametric EM and WH model. We first find that the proposed SNH model achieves a competitive or better performance in the case of exponential kernels. Remarkably, our model does a better job in capturing the delaying effect in the power-law kernel when compared to other models. An analysis of the above models based on the L1 error (fig \ref{Synthetic_data_1}) proves that the accuracy of kernel estimation is higher in SNH model. From the convergence plot of the SNH model in figure  \ref{Synthetic_data_1}, we find that our model has the ability to minimise the negative log-likelihood and meet the ground value swiftly in the first few iterations.

\subsection{K-fold cross validation of real data}\label{B3}
Cross-validation is one of the most widely used method for evaluating learning algorithms. Ideally, we divide the dataset into training set, cross-validation set and test set, to optimize the parameters, evaluate each algorithm and finally test the successful algorithm with least error. However, when the data is scarce or limited we are left with fewer numbers of samples in the training set. As a solution to this problem, we use the k-fold cross validation method \cite{friedman2001elements} to test the performance of our model. In this method, we divide the dataset into k-groups and for each of these groups we split the training and test set to evaluate the score. The performance measure is the average of the evaluated scores of the k-groups, given as, 

\begin{equation}
CV(\Theta) = \dfrac{1}{K}\sum_{k=1}^{K} \mathcal{L}(\Theta)
\end{equation}

\begin{figure}
  \centering
  \includegraphics[width=\linewidth]{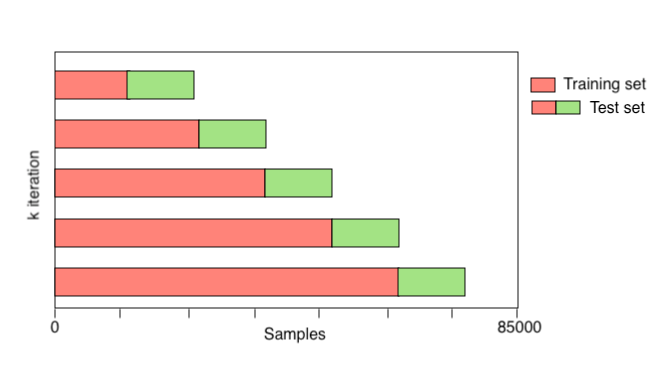}
  \caption{TimeSeriesSplit function on bivariate Bitcoin dataset (buy and sell data) with $N_T = 83574$ and number of splits = 5 }
  \label{cv_1}
\end{figure}

For the dataset in our experiment, we use the TimeSeriesSplit function provided by Scikit-learn \cite{scikit-learn}. Unlike non-time series data where the data are randomly split, this function divides the dataset along with the sequence and successive training sets are supersets of those that come before them. Due to the dependence on history in Hawkes processes, we modify the split function in order to evaluate the negative log-likelihood collectively on training and test samples (rather than on just test samples). The figure \ref{cv_1} demonstrates the time-series cross-validation split on bivariate Bitcoin data, for K = 5 groups. Training sets are of sizes ($N_T$)  $13928, 27857, 41786, 55715, 69644$ and their corresponding test sets are of sizes  ($N_T$)  $27857, 41786, 55715, 69644, 83573$. The estimated score, i.e the negative log-likelihood values for SNH, EM and WH models are $-21963.31$, $-19853.862$ and $147752.28$ respectively. 

\subsection{Additional analysis on synthetic data}\label{B4}

\begin{figure}
  \centering
  \begin{subfigure}[b]{0.45\linewidth}
    \includegraphics[width=\linewidth]{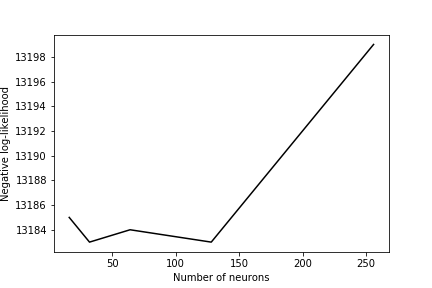}
    \caption{}
  \end{subfigure}
  \begin{subfigure}[b]{0.45\linewidth}
    \includegraphics[width=\linewidth]{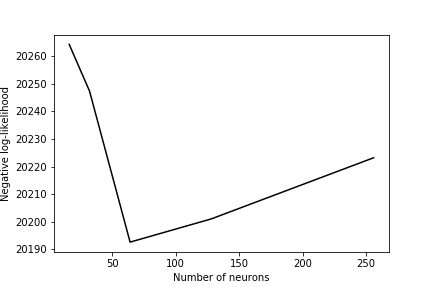}
    \caption{}
  \end{subfigure}
  \caption{Plot of estimated negative log-likelihood for varied number of neurons \\ (a) The case of exponential kernels  (b) The case of rectangular kernels }
  \label{Neuron_vs_likelihood}
\end{figure}

We extend the analysis on univariate Hawkes process described in Section \ref{univariate} to study the effect of hyper-parameter choices for SNH. We first study the impact on the performance of SNH model with varied number of neurons. The figure \ref{Neuron_vs_likelihood} shows the estimated negative log-likelihood values with increasing number of neurons used in the SNH for the exponential form of kernel. We find, as would be expected,  that fewer number of neurons are sufficient to achieve convergence. Next, we perform similar analysis on the rectangular kernel described in \ref{univariate} and the results are demonstrated in Figure \ref{Neuron_vs_likelihood}. In this case, it is evident that optimum performance is achieved by using neurons in range $32 ~\text{to} ~128$ in the SNH architecture.

Next, we investigate the performance of the SNH model based on different choices of learning rates using the rectangular kernel described in \ref{univariate}, for a fixed number of epochs. We find that using higher learning rates for the outer layer in comparison to the inner layer helps in faster convergence of the results. 

\begin{table}
  \label{table 3}
  \centering
  \begin{tabular}{ll}
    \toprule
    Lr of outer layer & Neg loglik  \\
    \midrule
     0.0001&37113.09  \\
     0.005&20244.66 \\
     0.001&20562.68 \\
     0.05& 20765.67\\
     0.01&20201.20\\
    \bottomrule
  \end{tabular}
  \caption{Estimated negative log-likelihood for different learning rates of the outer layer of SNH model (inner layer = $5 \times 10^{-4}$)}
\end{table}

\begin{table}
  \label{table 4}
  \centering
  \begin{tabular}{ll}
    \toprule
    Lr of inner layer & Neg loglik  \\
    \midrule
    0.00001 & 21369.25\\
    0.00005 & 20483.83 \\
    0.0005&20350.01 \\
     0.0001&20321.78 \\
     0.005 &20587.84\\
     0.01 & 22346.72\\
    \bottomrule
  \end{tabular}
  \caption{Estimated negative log-likelihood for different learning rates of the inner layer of SNH model (outer layer = $10^{-2}$)}
\end{table}
\end{document}